\documentclass[letterpaper,journal]{IEEEtran}
\usepackage{amsmath,amsfonts}
\usepackage{algorithmic}
\usepackage{array}
\usepackage[caption=false,font=small,labelfont=sf,textfont=sf]{subfig}
\usepackage{textcomp}
\usepackage{stfloats}
\usepackage{url}
\usepackage{verbatim}
\usepackage{graphicx} 
\usepackage{tabularx}
\usepackage{booktabs}
\usepackage{multirow} 
\usepackage[T1]{fontenc}
\usepackage{booktabs}
\usepackage{bigstrut}
\usepackage{amsmath}
\usepackage{graphicx}
\usepackage{epstopdf}
\usepackage{CJKutf8}
\usepackage{booktabs}    
\usepackage[section]{placeins}
\usepackage{float}
\usepackage[table]{xcolor}
\usepackage{makecell}  
\usepackage{color}
\usepackage{gensymb}

\usepackage[ruled,linesnumbered]{algorithm2e}

\usepackage[numbers,sort&compress]{natbib}
\def\BibTeX{{\rm B\kern-.05em{\sc i\kern-.025em b}\kern-.08em
		T\kern-.1667em\lower.7ex\hbox{E}\kern-.125emX}}
\DeclareSubrefFormat{myparens}{#1(#2)}

\begin{document}
	
	\begin{CJK}{UTF8}{gbsn}
		\title{Simultaneous Enhancement and Noise Suppression under Complex Illumination Conditions}
		
		\author{Jing Tao, You Li, Banglei Guan, Yang Shang, and Qifeng Yu 
			\thanks{Manuscript received 20 February 2024; revised 22 May 2024;
				accepted 2 July 2024. Date of publication 16 August 2024; 
				date of current version 27 August 2024. This work was supported 
				in part by the Hunan Provincial Natural Science Foundation for 
				Excellent Young Scholars under Grant 2023JJ20045, in part by the 
				Science Foundation under Grant KY0505072204, and in part by the 
				National Natural Science Foundation of China under Grant 12372189. 
				The Associate Editor coordinating the review process was 
				Dr. Kunpeng Zhu. (Corresponding authors: Banglei Guan; Yang Shang.)}
			\thanks{Jing Tao is with the College of Aerospace Science, National University of Defense Technology, Changsha 410000, China (e-mail :tao.j@nudt.edu.cn).
				
				You Li is with the Key Laboratory of Human Factors Engineering, China Astronaut Research and Training Center, Beijing 100094, China (e-mail: liyou@nudt.edu.cn)
				
				Banglei Guan, Yang Shang and Qifeng Yu are with the College of Aerospace Science and Engineering, National University of Defense Technology, Changsha 410000, China (e-mail: guanbanglei12@nudt.edu.cn; yangshang1977@nudt.edu.cn).}}			
		
		\markboth{IEEE TRANSACTIONS ON INSTRUMENTATION AND MEASUREMENT, VOL.**.2024}%
		{Shell \MakeLowercase{\textit{et al.}}: A Sample Article Using IEEEtran.cls for IEEE Journals}

		\maketitle
		
		\begin{abstract}
			Under challenging light conditions, captured images often suffer from various degradations, leading to a decline in the performance of vision-based applications. Although numerous methods have been proposed to enhance image quality, they either significantly amplify inherent noise or are only effective under specific illumination conditions. To address these issues, we propose a novel framework for simultaneous enhancement and noise suppression under complex illumination conditions.
			Firstly, a gradient-domain weighted guided filter (GDWGIF) is employed to accurately estimate illumination and improve image quality. 
			Next, the Retinex model is applied to decompose the captured image into separate illumination and reflection layers. These layers undergo parallel processing, with the illumination layer being corrected to optimize lighting conditions and the reflection layer enhanced to improve image quality.
			Finally, the dynamic range of the image is optimized through multi-exposure fusion and a linear stretching strategy. 
			The proposed method is evaluated on real-world datasets obtained from practical applications. Experimental results demonstrate that our proposed method achieves better performance compared to state-of-the-art methods in both contrast enhancement and noise suppression.
		\end{abstract}
		
		\begin{IEEEkeywords}
			Image enhancement, noise suppression, guided filter, Retinex model, multi-exposure.
		\end{IEEEkeywords}

		\section{Introduction}
		\IEEEPARstart{I}{mages} captured under challenging conditions often suffer from poor quality, significantly impacting the accuracy of vision-based measurement instrumentation. These images frequently exhibit various degradations, including limited visibility, low contrast, and noticeable noise \cite{2022Low,GACA,CovDenseSNN}.
		To address these issues, researchers have attempted to improve image quality using professional devices and advanced photographic techniques \cite{High-speed,guan2023minimal,Decoding,guan2022monitoring,huang2022}. However, these efforts remain limited in certain scenarios, such as space environments. In space, direct sunlight and stellar occlusion introduce additional interferences, including backlighting, uneven illumination, and reflections from spacecraft components. Consequently, the reduced contrast of astronaut identification features poses significant challenges for detection and recognition tasks. Therefore, enhancing visual quality in complex scenes remains a critical objective.
		
		\begin{figure}[t]
			\centering
			\includegraphics[scale=0.315]{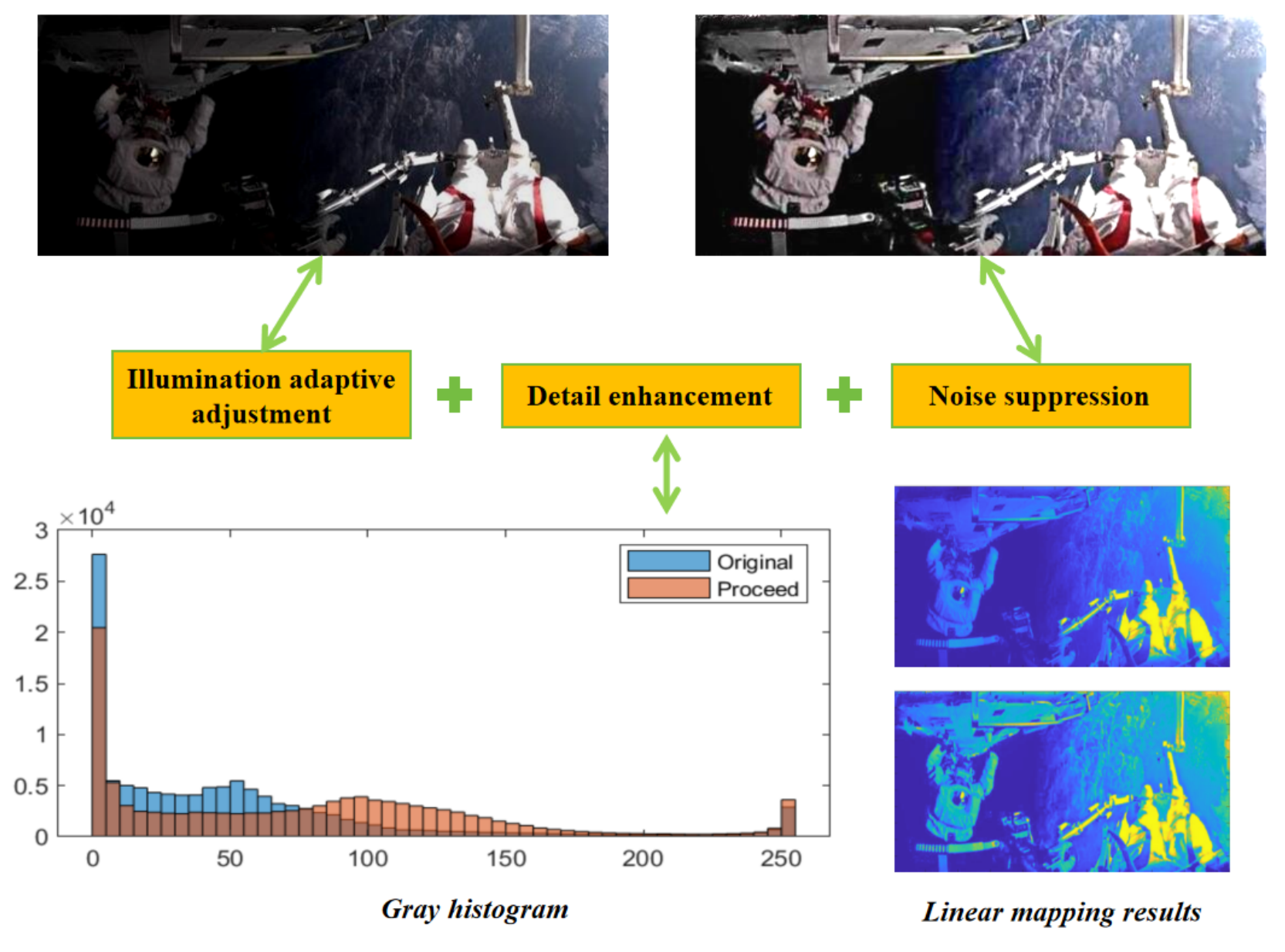}
			\caption{An example of our proposed adjustment framework under complex illumination conditions. The gray distribution map and linear pixel mapping map are utilized to demonstrate the changes before and after adjustment.}
			\label{fig:taster}
		\end{figure}
		
		Addressing these challenges is a complex undertaking due to the presence of uneven illumination and significant noise. For one thing, accurate illumination estimation and appropriate correction methods are essential to balance scene brightness. For another, images captured under low-light conditions often exhibit high levels of noise, which may be comparable in intensity to edges and textures. Therefore, effectively suppressing noise while enhancing fine details is essential. In this paper, methods designed to enhance image quality under complex illumination conditions must address problems such as limited visibility, low contrast, and prominent noise.
		
		Traditional image enhancement methods typically manipulate pixel values directly to improve contrast and brightness. Although methods such as histogram equalization \cite{HE,2010Adaptive,2022Histogram} and gamma correction \cite{2020comprehensive,2021Low} can enhance visibility to some extent, they lack a physical interpretation and may cause over- or under-enhancement, leading to suboptimal results \cite{2021SCENS}. In contrast, the Retinex theory has emerged as an alternative approach, offering a theoretical understanding of the human visual system and demonstrating promising results in image enhancement \cite{1986retinex,2017retinex,2020star,2020A}. A key aspect of this theory involves simultaneously estimating illumination and reflectance, allowing researchers to effectively separate the influence of lighting conditions from the intrinsic characteristics of captured objects or scenes \cite{2020comprehensive,2021Total}.
		However, some Retinex-based approaches may produce halo artifacts around edges, resulting in unusual image features \cite{2014A}.
		To address this, Guo \emph{et al.} \cite{3} incorporated structural perception as a constraint for the illumination component of the model. However, the lack of constraints on reflection made it less resistant to noise. Subsequently, the approach presented in \cite{4} integrated noise-related concerns into the standard Retinex model, resulting in a more robust solution. Nonetheless, this approach may lead to a loss of local contrast under complex light conditions, thereby limiting its applicability. 
		
		Recently, the field of image enhancement has seen a significant surge of interest in machine learning-based approaches \cite{2020Learning,deep2021,Network}. Researchers have explored the capabilities of deep learning models to improve the visual quality and perception of images. For instance, Zhang \emph{et al.} \cite{Underexposed} introduced a deep learning approach that focuses on detail recovery and enhances visual perception in underexposed images through automatic learning. Similarly, Li \emph{et al.} \cite{LDNet} proposed an efficient network named LDNet, which addresses both low-light enhancement and denoising tasks concurrently.
		However, these machine learning-based methods also have certain limitations. They often overlook the underlying physical interpretation, resulting in unnatural correction outcomes. Additionally, the performance of these deep learning models heavily depends on the quality of training datasets. Obtaining specialized datasets for particular scenes can pose challenges, potentially limiting the effectiveness of these approaches \cite{2021SCENS,2021Low}.
		
		\begin{figure*}[htp]
			\centering 
			\includegraphics[width=0.96\textwidth,height=0.37\textwidth]{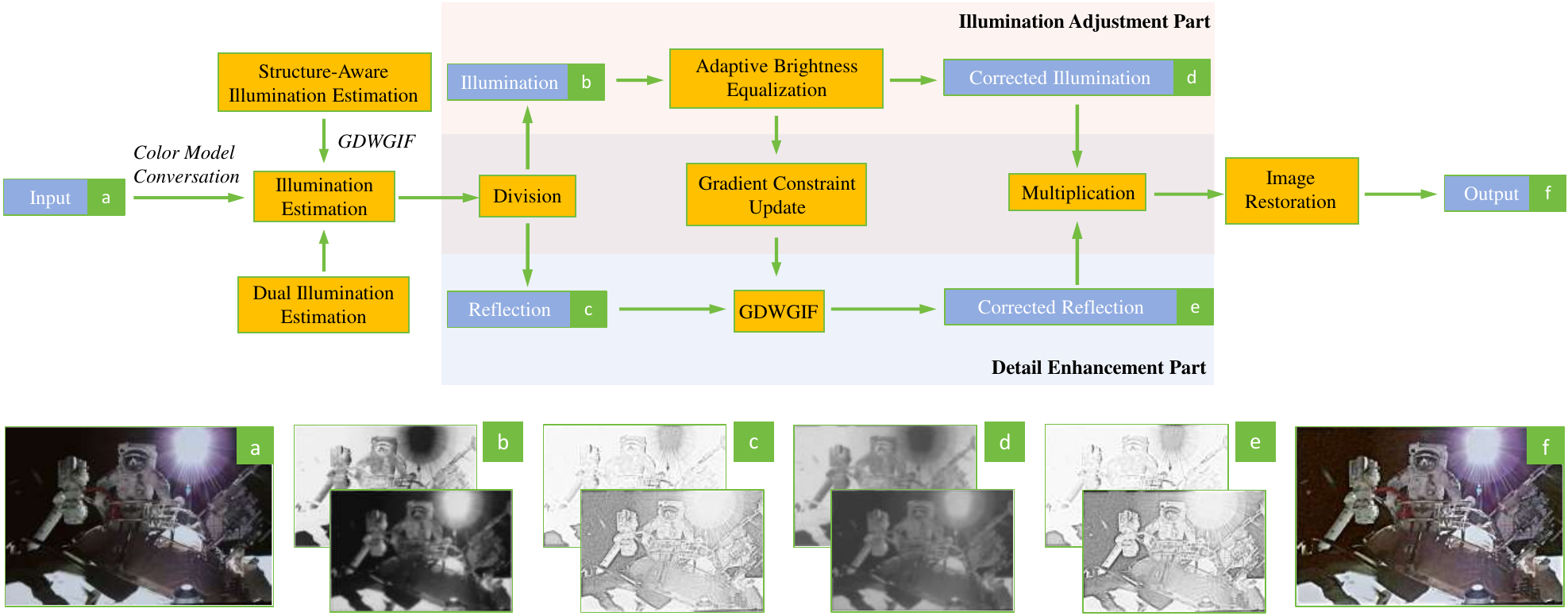}
			\caption{The schematic diagram illustrates the proposed framework for simultaneous image enhancement and noise suppression. Orange boxes: methods. Blue boxes: input or output images of each stage.}
			\label{fig:Overview}
		\end{figure*}
		
		In this paper, we propose a novel image enhancement framework based on an improved guided filter. Our framework aims to simultaneously improve image quality and suppress noise under challenging illumination conditions.
		By capturing valuable illumination and reflectance while eliminating extraneous interference, the method adeptly compensates for brightness variations, enhances image details, and reduces noise.
		To demonstrate the efficacy of our proposed framework, visual results are presented in Fig.\ref{fig:taster}, showcasing significant enhancements in image quality. Remarkably, our framework effectively preserves overall detail information even in extreme lighting conditions,  while enhancing visibility in darker regions by revealing finer details and reducing noise.
		The main contributions of this paper can be summarized as follows:
		\begin{itemize}
			\item[$\bullet$] This paper introduces a novel framework for enhancing images captured under complex lighting conditions. It achieves simultaneous brightness compensation, detail enhancement, and noise suppression by effectively capturing essential illumination and reflectance while eliminating unwanted interference. 
			
			\item[$\bullet$] To suppress noise in images, the proposed framework incorporates a novel gradient-domain weighted guided filter (GDWGIF). This filter employs an edge-aware gradient strategy and an adaptive filtering window, enabling the optimization of the high-frequency signal within the image.
			
			\item[$\bullet$] To address uneven illumination in images, the proposed framework introduces an adaptive brightness equalization strategy. This strategy integrates adaptive gamma transform and dual-illumination fusion techniques, ensuring superior image quality even in complex lighting conditions.
		\end{itemize}
		
		The rest of this paper is structured as follows. Section II introduces the algorithm details of the proposed framework. Section III conducts a comparative analysis between the proposed method and other state-of-the-art methods. Finally, Section IV presents the conclusion of this paper.
		
		\section{Proposed methodology}
		The schematic diagram in Fig.\ref{fig:Overview} illustrates the proposed framework for simultaneous image enhancement and noise suppression. This framework consists of three primary stages: illumination estimation, adaptive correction, and image restoration. A crucial algorithm, the gradient-domain weighted guided filter (GDWGIF), is integrated throughout the framework to facilitate these processes.
		
		\begin{figure}[t]
			\centering
			\includegraphics[scale=0.41]{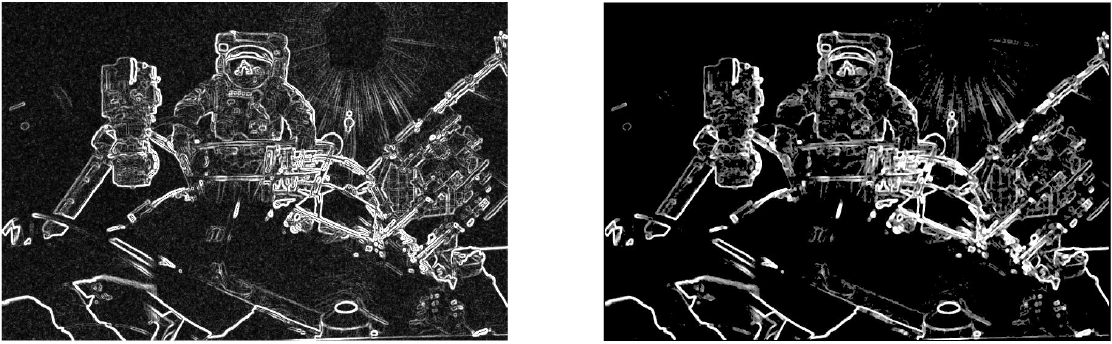}
			\caption{Comparison of gradient map. Left: convolution gradient map. Right: edge-sensing gradient map.}
			\label{fig:gradient_maps} 
		\end{figure}
		
		\subsection{Gradient-Domain Weighted Guided Filter}
		Guided filter (GIF) \cite{GIF} is renowned for its exceptional filtering performance and high efficiency. It establishes a local linear model between the filtering output $z$, and the guidance image $I$ within a local window ${\Omega _k}$ centered at pixel $k$:
		\begin{equation}
			{z_i} = ({a_k}{I_i} + {b_k}),\forall i \in {\Omega _k}
			\label{eq:zi}
		\end{equation}
		where $({a_k},{b_k})$ represent constant linear coefficients within ${\Omega _k}$.
		
		The cost function in the window is
		\begin{equation}
			E = \min \sum\limits_{i \in {\Omega _k}} {({{({a_k}{I_i} + {b_k} - {q_i})}^2}}  + \lambda {a_k}^2)
			\label{eq:GIF_min}
		\end{equation}
		which minimizes the difference between the output $z$ and the input $q$. Here, $\lambda $ is a regularization parameter preventing ${a_k}$ from being too large.
		However, due to its fixed nature, $\lambda $ inevitably leads to artifacts around strong edges \cite{WGIF,AIGIF,SKWGIF,WM}. To overcome this drawback, we propose a novel filter in this section, aimed at enhancing GIF performance. This method combines the edge-aware gradient strategy with the adaptive window strategy to achieve improved filtering efficacy.
		
		\vspace{0.1cm}
		\emph{1) Edge-sensing gradient strategy:} 
		Template convolution is a widely employed technique for extracting gradient information ($g$) from images. However, it may introduce additional noise when applied to noisy images, as depicted in Fig.\ref{fig:gradient_maps}. Hence, incorporating gradient suppression during the gradient extraction process is essential.
		
		Firstly, local variance is employed to classify edges into two distinct categories: weak edges (${g_w}$) and strong edges (${g_s}$). Here, ${{\sigma _{g,\xi }}({g_k})}$ denotes the local variance of gradient values within a window of radius $\xi $ at pixel $k$. ${mea{n_{{\sigma _{g,\xi }}({g_k})}}}$ denotes the mean of the local variances within the window. The details are as follows:
		\begin{equation}
			{\left\{ {\begin{array}{*{20}{c}}
						{g \in {g_w}}&{\rho  < Threshold}\\
						{g \in {g_s}}&{else}
				\end{array}} \right.}
			\label{eq:categorizing}
		\end{equation}
		\begin{equation}
			\rho  = \left| {\frac{{{\sigma _{g,\xi }}({g_k})}}{{mea{n_{{\sigma _{g,\xi }}({g_k})}}}} - 1} \right|
			\label{eq:frac}
		\end{equation}
		
		Subsequently, threshold segmentation and wavelet filtering are individually applied to each set to effectively distinguish noise from edges, resulting in the processed weak edge set ($g'_w$) and the processed strong edge set ($g'_s$). These sets are then merged to generate an edge-aware gradient map ($ g'$).
		\begin{equation}
			g' = {g'_w} + {g'_s}
			\label{eq:g}
		\end{equation}
		\begin{figure}[t]
			\centering
			\includegraphics[scale=0.34]{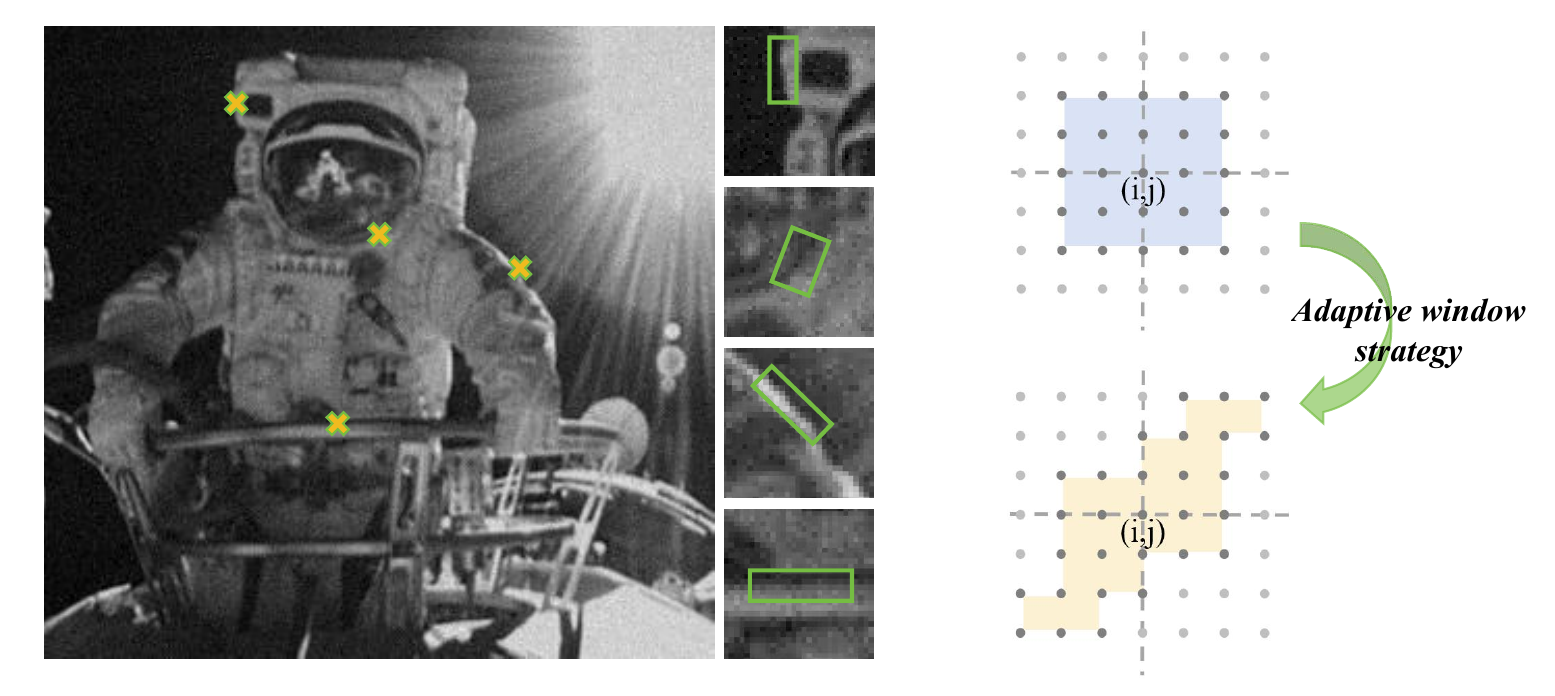}
			\caption{Adaptive window results. Left: the adaptive shape of the partial pixel window. Right: the change of pixels within the window.}
			\label{fig:adaptive_window}
		\end{figure}	
		
		\vspace{0.1cm}
		\emph{2) Adaptive window strategy:} 
		Conventional filtering methods typically employ rectangular windows, which may lead to edge information loss due to the anisotropic characteristics of edge pixels \cite{2013Local}. To address this issue, the following section introduces an adaptive window-shape filter designed to enhance noise reduction near edges.
		
		To improve algorithm efficiency, only pixels with non-90\degree gradient angles at the edge points are considered. Firstly, the horizontal and vertical gradients of the image, denoted as ${g_x}$ and ${g_y}$, are calculated. Then, the aspect ratio is calculated as ${\tau ^0} = \left| {({g_x} + 1)/({g_y} + 1)} \right|$ based on the pixel gradient. The initial window is determined based on ${\tau^0}$.
		
		\begin{equation}
			{\begin{array}{*{20}{c}}
					{{W^0} = {\tau ^0} \cdot r,}&{{H^0} = \frac{{r \times r}}{W}}
			\end{array}}
			\label{eq:size}
		\end{equation}
		\begin{equation}
			{\theta ^0} = \arctan (\frac{{{g_y}}}{{{g_x}}})
			\label{eq:angle}
		\end{equation}	
		the dimensions of the initial window are represented by ${W^0}$ for length and ${H^0}$ for width. The initial window's normal direction is denoted by ${\theta ^0}$, and $r$ represents a fixed coefficient.
		
		The window properties are updated by calculating the sum of pixel gradients for a set of $N$ pixels within the window range ${\Omega}$, along with the sum of their absolute values. Here, $(i,j)$ denotes the coordinates of the central pixel within the window, while $m$ and $n$ represent the horizontal and vertical coordinates of the pixels within the window relative to the central pixel.
		
		The sums of the horizontal and vertical gradients are denoted as ${\bar g_x}$ and ${\bar g_y}$, respectively:
		
		\begin{equation}
			\left\{ \begin{array}{l}
				{{\bar g}_x}(i,j) = \sum\limits_{(i + m,j + n) \in \Omega} {{g_x}} (i + m,j + n)\\
				{{\bar g}_y}(i,j) = \sum\limits_{(i + m,j + n) \in \Omega} {{g_y}} (i + m,j + n)
			\end{array} \right. 
			\label{eq:sum_gradient}
		\end{equation}
		the sums of the absolute values of the horizontal and vertical gradients are denoted as ${{{\tilde g}_x}}$ and ${{{\tilde g}_y}}$, respectively:
		
		\begin{equation}
			\left\{ {\begin{array}{*{20}{l}}
					{{{\tilde g}_x}(i,j) = \sum\limits_{(i + m,j + n) \in \Omega} {\left| {{g_x}(i + m,j + n)} \right|} }\\
					{{{\tilde g}_y}(i,j) = \sum\limits_{(i + m,j + n) \in \Omega} {\left| {{g_y}(i + m,j + n)} \right|} }
			\end{array}} \right. 
			\label{eq:ab_gradient}
		\end{equation}
		
		Then, $ {\tau^0} $ undergoes iterative updates based on the results of Eq.\eqref{eq:sum_gradient} to adjust the window size in Eq.\eqref{eq:size}. Simultaneously, results derived from Eq.\eqref{eq:ab_gradient} replace $g$ in Eq.\eqref{eq:angle} to refine the window direction. This iterative process continues until convergence is achieved, resulting in the final adaptive window, as shown in Fig.\ref{fig:adaptive_window}. Within the adaptive window range, either a custom Gaussian filter or a median filter is employed to effectively suppress noise near edges.
		
		\begin{figure*}[t]
			\centering		
			\setlength{\abovecaptionskip}{0.3cm}
			\subfloat[Input]{
				\includegraphics[scale=0.55]{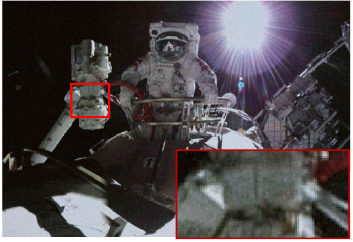}}
			\hspace{0.025cm}
			\subfloat[GIF \cite{GIF}]{	
				\includegraphics[scale=0.55]{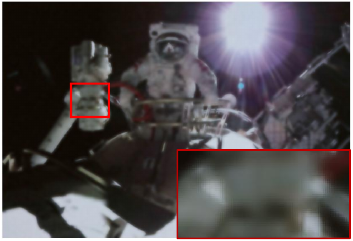}}
			\hspace{0.025cm}	
			\subfloat[WGIF \cite{WGIF}]{	
				\includegraphics[scale=0.55]{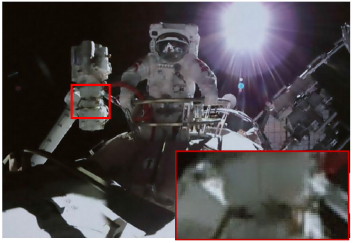}}
			\hspace{0.025cm}    
			\subfloat[EGIF \cite{EGIF}]{	
				\includegraphics[scale=0.55]{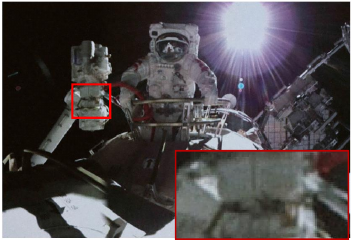}}
			\hspace{0.025cm}	
			\subfloat[GDWGIF]{	
				\includegraphics[scale=0.55]{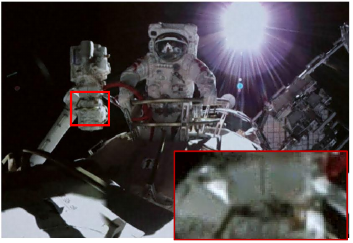}}
			\caption{Comparison of four filters, the regularization parameter $\lambda $ is 0.1 and the window size is 7 $\times$ 7. (a) The input image. (b) GIF \cite{GIF} filtered image. (c) WGIF \cite{WGIF} filtered image. (d) EGIF \cite{EGIF} filtered image. (e) The proposed GDWGIF filtered image.}
			\label{fig:Comparison_GIF}
		\end{figure*}
		
		Building on the above analysis, we combine the existing improved guided filter to develop the final enhanced filter \cite {GDGIF,WGIF}. An edge-aware weight, denoted as ${\hat T_I(k)}$, is introduced to improve the regularization term of the filter:
		\begin{equation}
			\begin{array}{*{20}{c}}
				{{{\hat T}_I}(k) = \frac{1}{N}\sum\limits_{p = 1}^N {\frac{{\chi (k) + \varepsilon }}{{\chi (k) + \varepsilon }}} ,}&{\chi (k) = {\varphi _{I,3}}{\varphi _{I,\xi }}g'}
			\end{array}
			\label{eq:weight}
		\end{equation}
		here, $\varepsilon $ is a small constant defined as ${\left( {0.001 \times {\cal L}} \right)^2}$, where ${\cal L}$ is the dynamic range of the input image. ${\varphi _{I,3}} = {\sigma _{I,3}}({g'_k})/mea{n_{{\sigma _{I,3}}}}$ represents the coefficient of variation of gradient information with a window radius of 3 centred at pixel $k$, and ${\varphi _{I,\xi }} = {\sigma _{I,\xi }}(k)/mea{n_{{\sigma _{I,\xi }}}}$ denotes the coefficient of variation at pixel $k$ with a radius of ${\xi}$. $g'$ denotes the gradient information obtained by the above method. So, the cost function is changed as follows:
		\begin{equation}
			\small
			E = \min \sum\limits_{i \in {\Omega _k}} {({{({a_k} \times {I_i} + {b_k} - {q_i})}^2}}  + \frac{\lambda }{{{{\hat T}_I}(k)}} \times {({a_k} - {\psi _k})^2})
			\label{eq:GDWGIF_min}
		\end{equation}
		where ${\psi _k}$ is defined as:
		\begin{equation}
			{\psi _k} = 1 - \frac{1}{{1 + {e^{\eta (\chi (k) - {\mu _{\chi ,\infty }})}}}}
			\label{eq:psi}
		\end{equation}
		here, ${{\mu _{\chi ,\infty }}}$ denotes the mean value of all ${\chi (k)}$, and $\eta $ is calculated as $4/({\mu _{\chi ,\infty }} - \min (\chi (k)))$. Combining  Eq.\eqref{eq:zi} and Eq.\eqref{eq:GDWGIF_min}, we can get the final result $z$.
		
		As shown in Fig.\ref{fig:Comparison_GIF}, the proposed filter significantly enhances image clarity and overall quality compared to other approaches. The complete procedure of the GDWGIF is detailed in Algorithm \ref{alg:1}.
		
		\begin{algorithm}[htbp]
			\caption{Gradient-domain weighted guided filter (GDWGIF) }
			\label{alg:1}	
			\SetKwData{Left}{left}\SetKwData{This}{this}\SetKwData{Up}{up}
			\SetKwFunction{Union}{Union}\SetKwFunction{FindCompress}{FindCompress}
			\SetKwInOut{Input}{input}\SetKwInOut{Output}{output}
			\Input{Image $q$, $\lambda $ and window radius $\xi $}  
			\Output{Result $z$}    
			\BlankLine
			parameters setting: $Threshold = 0.2$, $r = 5$\\
			initialize guided image $I = q$\\ 
			compute optimized gradient $g'$ via Eq.\eqref{eq:categorizing} to Eq.\eqref{eq:g}\\
			compute $z$ via Eq.\eqref{eq:zi} and Eq.\eqref{eq:weight} to Eq.\eqref{eq:psi}\\
			\For{each row in gradient map}
			{\For{each column in gradient map}{
					\If{$magnitude \ne 0$ and $angle \ne 90^\circ $ }
					{initialize the window via Eq.\eqref{eq:size} and Eq.\eqref{eq:angle}\\
						\While{not coveraged}
						{
							compute $\bar g$ and $\tilde g$ via Eq.\eqref{eq:sum_gradient} and Eq.\eqref{eq:ab_gradient}\\
							update ${\tau ^{(k + 1)}} = \left| {({{\bar g}_x} + 1)/({{\bar g}_y} + 1)} \right|$\\
							update ${l^{(k + 1)}}$ and ${w^{(k + 1)}}$ via Eq.\eqref{eq:size}\\
							update ${\theta ^{(k + 1)}}$ using ${g} = {\tilde g}$ in Eq.\eqref{eq:angle}
						}
						optimal $z$ via adaptive window filter
			}	}}
		\end{algorithm}
		
		\subsection{Illumination Estimation}
		\emph{1) Structure-aware illumination estimation:} 
		To address potential overflow issues arising from extremely low illumination values during the image recovery process, a method is proposed for computing the initial illumination component $L$. This involves selecting the maximum value from the RGB color channels in the input image $q$ at each pixel, which is expressed as:
		\begin{equation}
			{L_k} = \max q_k^c,\forall c \in \{ r,g,b\} 
			\label{eq:initial_L}
		\end{equation}
		here, $q_k^c$ represents the color channel $c$ at pixel $k$. 
		
		To refine the accuracy of light component estimation, a hybrid approach combining coarse and fine methods is adopted. Specifically, a multi-scale GDWGIF filtering approach is utilized to capture both global and local characteristics of the estimated illumination value. The illumination component of the image is extracted at three different scales, with corresponding weight coefficients assigned to each scale.
		To further refine the illumination estimation and reduce excessive enhancement caused by residual texture information, a multi-objective optimization function is integrated to obtain the fine illumination estimation, denoted as ${\hat L}$ \cite{7}.
		
		As depicted in Fig.\ref{fig:Overview}b, the proposed illumination estimation method adeptly removes extraneous texture details from the initial illumination while preserving essential structural aspects. This facilitates the subsequent illumination correction process. Additionally, in the subsequent dual illumination estimation, this method effectively estimates the illumination in the dual-channel image.
		
		\vspace{0.1cm}
		\emph{2) Dual illumination estimation:} 
		Given the complexities of illumination in space environments, relying solely on a single estimation method may prove insufficient. Therefore, the concept of double illumination estimation is introduced \cite{5}. Specifically, the image is divided into positive and negative channels for concurrent processing. In this paper, the negative image is derived as ${{q}_{inv}} = 1 - q$, and the corresponding refined illumination ${{\hat L}_{inv}}$ is estimated. In subsequent discussions, the image set of these two channels, obtained through structure-aware illumination estimation, is collectively denoted as $\ell $, as illustrated in Fig.\ref{fig:Overview}b. This approach enables the adjustment of brightness under various illumination conditions.
		
		\begin{figure}[t]
			\centering	\includegraphics[scale=0.42]{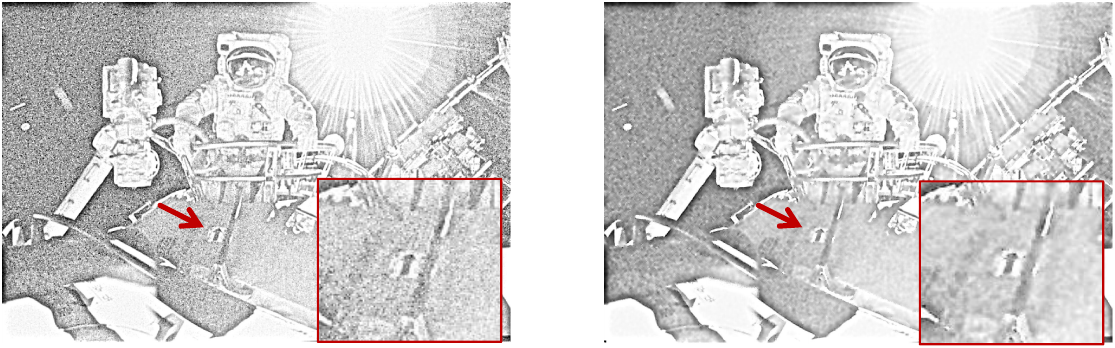}
			\caption{Comparison of reflection layers correction results. Left: the image before filtering. Right: the image after filtering.}
			\label{fig:denoise}
		\end{figure}
		
		\subsection{Adaptive Correction}
		\emph{1) Illumination layer correction:}
		To reveal obscured details in highlighted scenes and enhance target information in dark scenes by increasing brightness, gamma correction is applied. The corrected illumination layer ${\hat L'}$, ranging from 0 to 1, is computed as follows:
		\begin{equation}
			{\hat L_k'} = {[{\hat L_k}]^\gamma }
			\label{eq:ok}
		\end{equation}
		here, $\gamma $ is a control parameter.
		
		It has been observed that designing $\gamma $ solely based on pixels may result in unnatural transition areas in the correction results \cite{2021Weak}. To address this limitation, the local average of pixels replaces individual pixel values, balancing local and global information. This modification leads to a more comprehensive formulation of $\gamma $, enhancing its adaptability and performance in correction tasks.
		
		\begin{equation}
			\begin{array}{*{20}{c}}
				{\gamma  = {{\rm{m}}^{[2 \times \mu (k) - 1]}},}&{{\rm{m}} = \alpha  + \mu (k)}
			\end{array}
			\label{eq:gamma}
		\end{equation}
		where $\alpha$ denotes the gamma correction adjustment factor, and ${\mu (k)}$ denotes the local mean of pixel $k$. Through this process, the image quality of the illumination layer can be effectively corrected.
		
		\vspace{0.1cm}
		\emph{2) Reflection layer correction:}
		According to the Retinex theory, an image is generated by the reflection or emission of light from objects within a scene, which is then captured by the camera. Consequently, the image reflection layer $R$ can be derived from the obtained image brightness layer ${\hat L}$ using the following equation:
		\begin{equation}
			{R_k} = {L_k}/(\hat L_k + \tau )
			\label{eq:R}
		\end{equation}
		here, $\tau $ denotes a constant regularization term introduced to enhance image details. Similarly, the expansion of the reflection layer into two channels, as elaborated in section II.B, yields the set $\smallint $, which corresponds to $\ell $.
		
		The reflection layer, comprising edge information, detail information, and noise, encapsulates the high-frequency elements of the image. Thus, it is necessary to apply GDWGIF to obtain the processed reflection layer ${R'}$, simultaneously achieving noise suppression and edge enhancement. Importantly, variations in illumination can significantly influence the image's gradient information. Thus, the gradient constraint of the filter is updated using the image gradient information under the corrected illumination.
		
		Fig.\ref{fig:denoise} shows an enlarged view of the partial results obtained from Fig.\ref{fig:Overview}c and Fig.\ref{fig:Overview}e. As illustrated in Fig.\ref{fig:denoise}, the filtering process effectively suppresses image noise while enhancing details, as highlighted in the figures.
		
		\subsection{Image Restorationt}
		\emph{1) Multi-exposure image fusion:} 
		This section requires the fusion of results processed through multiple exposure channels, including the integration of the original input image to retain the normally exposed areas. To achieve this, multi-exposure image fusion is applied to the three images: the under-exposed fusion result ${{{q'}_f}}$, the over-exposed fusion result ${{{q'}_r}}$, and the original input image $q$. By applying multi-exposure fusion processing to the image sequence $\left\{ {{{q'}_f},{{q'}_r},q} \right\}$, well-exposed final results can be achieved \cite{2010Exposure,2020A}. The resulting image is then multiplied by its color factor and transformed into a color image.
		
		\begin{algorithm}[t]
			\caption{The proposed framework }
			\label{alg:2}	
			\SetKwData{Left}{left}\SetKwData{This}{this}\SetKwData{Up}{up}
			\SetKwFunction{Union}{Union}\SetKwFunction{FindCompress}{FindCompress}
			\SetKwInOut{Input}{input}\SetKwInOut{Output}{output}
			\Input{Image $q$, $\lambda $, window radius $\xi $ and adjustment factor $\alpha$}  
			\Output{Result ${{\hat q}_{out}}$}    
			\BlankLine 
			initialize ${{q}_{inv}} = 1 - q$\\ 
			\eIf{$q$ is triple channel}
			{initialize  $L$ and ${L}_{inv}$ via Eq.\eqref{eq:initial_L}}
			{$L = q$ and ${{\rm{L}}_{inv}} = 1 - q$ }
			compute $\ell  = \{ \hat L,{{\hat L}_{inv}}\} $ via refined illumination estimation \\
			compute $\smallint  = \{ R,{R_{inv}}\} $ via Eq.\eqref{eq:R} \\
			\For{each element in $ \ell $}
			{compute ${\hat L'}$ via Eq.\eqref{eq:ok}}
			\For{each element in $ \smallint $}
			{
				update GDWGIF via Eq.\eqref{eq:g}\\ compute $R'$ via GDWGIF
			}
			compute ${{q'}_f} = \hat L' \cdot R'$ and ${{q'}_r} = {{\hat L'}_{inv}} \cdot {R'_{inv}}$\\
			compute ${q_{out}}$ via multi-exposure image fusion\\
			compute ${{\hat q}_{out}}$ via Eq.\eqref{eq:linear}
		\end{algorithm}
		
		\vspace{0.3cm}
		\emph{2) Grayscale linear stretching:} 
		To address the issue of localized variations in image gray values, a grayscale stretching function is employed to enhance the overall contrast of the brightness-adjusted image. The resulting corrected image ${q_{out}}$
		is then subjected to linear stretching to generate the final image ${{\hat q}_{out}}$.
		The equation for linear stretching is given by:
		\begin{equation}
			{{\hat q}_{{\rm{out}}}} = \frac{{{q_{{\rm{out}}}} - \min {q_{{\rm{out}}}}}}{{\max {q_{{\rm{out}}}} - \min {q_{{\rm{out}}}}}}
			\label{eq:linear}
		\end{equation}
		here, $\min $ and $\max $ denote the minimum and maximum pixel values in the image ${q_{out}}$, respectively. 
		
		As shown in Fig.\ref{fig:Overview}f, the implementation of our proposed algorithm extends the dynamic range of the image, resulting in enhanced overall visual performance. The complete framework for image enhancement is detailed in Algorithm \ref{alg:2}.
		
		\begin{figure}[t]
			\centering	\includegraphics[width=0.48\textwidth]{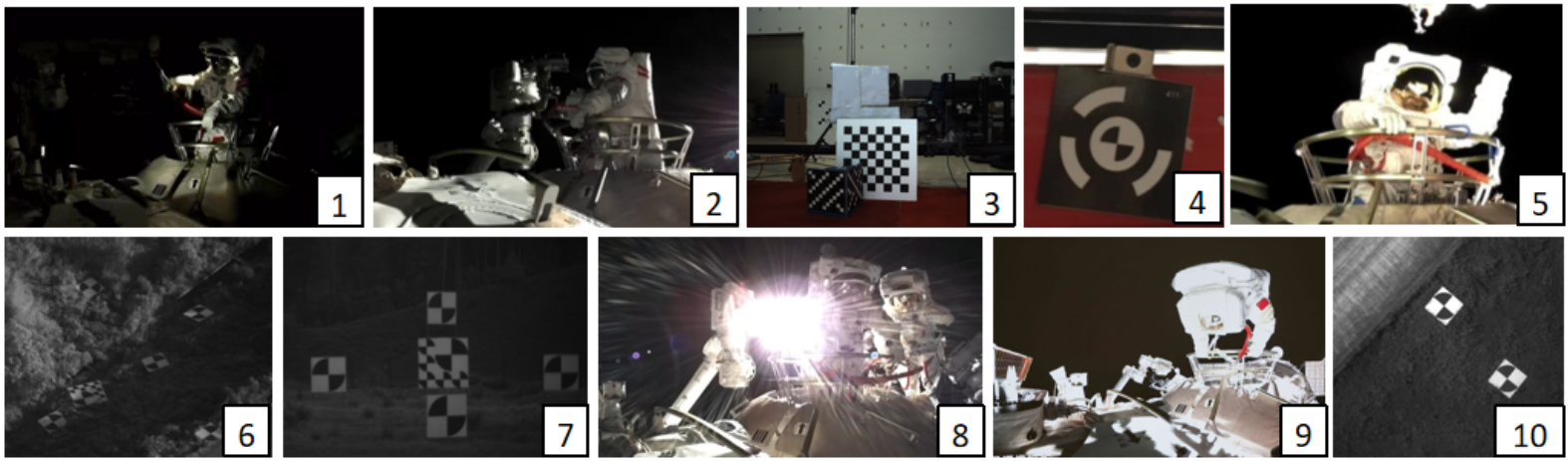}
			\caption{Some test images used in our experiments.}
			\label{dataset}
		\end{figure}
		
		\section{Experimental results}
		This section evaluates the performance of the proposed method. It begins by presenting the details of the implementation. Following this, a comparison is drawn between the proposed method and state-of-the-art light correction methods using both qualitative and quantitative evaluations. Noise suppression results are then presented. Finally, experiments related to diagonal marker extraction and other applications are discussed.
		\begin{figure*}[b]
			\centering
			\includegraphics[scale=0.56]{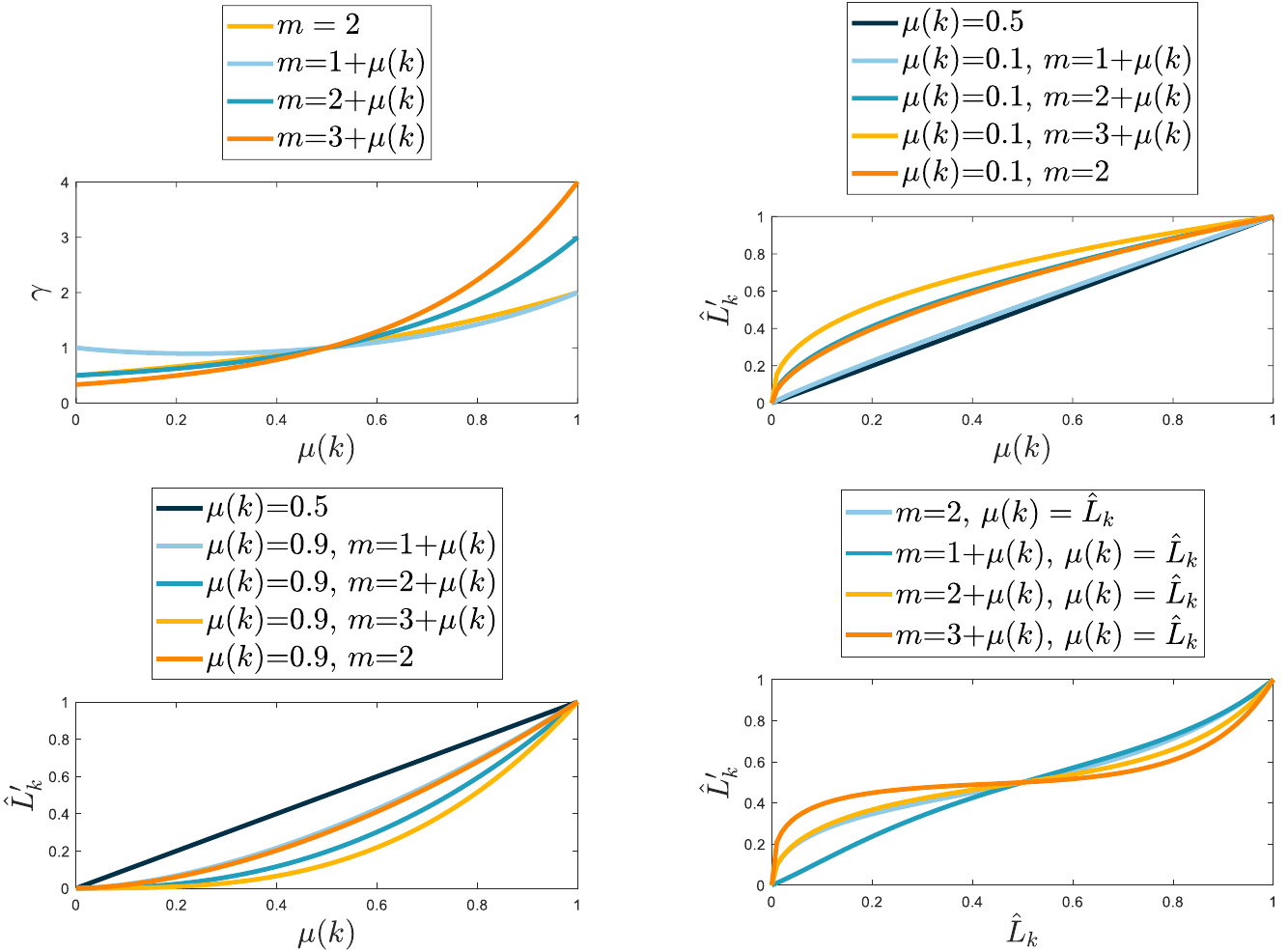}
			\caption{Change in the illumination curve map under various parameter selections.}
			\label{fig:Curve changes}
		\end{figure*}
		
		\subsection{Implementation Details}
		The proposed method is implemented using MATLAB R2022a on a Windows 11 OS, with an AMD Ryzen 7 5800U with Radeon Graphics CPU and 32 GB RAM.
		During the implementation, the parameters $r$ and $\lambda$ are set as 5 and 0.2, respectively. The \emph{Threshold} for gradient edge perception is set at 0.2. For the adaptive gamma correction part, selecting the parameter $\alpha $ in Eq.(\ref{eq:gamma}) is crucial to balance overall brightness and contrast. To observe the corresponding changes in the correction effect, we varied the parameter values to 1, 2, or 3, as illustrated in Fig.\ref{fig:Curve changes}. Considering the visual impact on the image, we opted for a moderate range of variation and set $\alpha $ to 2. 
		
		The performance is assessed qualitatively and quantitatively by comparing it with several conventional and state-of-the-art low-light image enhancement methods, including CLAHE \cite{1}, MARCR \cite{2}, LIME \cite{3}, SRLIME \cite{4}, DUAL \cite{5}, LIESD \cite{6}, and NPLIE \cite{7}. To evaluate our method's performance in the extreme space environment, two distinct datasets were curated. The first dataset comprises image sequences documenting activities performed by Chinese astronauts outside the cabin, generously provided by China Central Television (CCTV). The second dataset consists of images specifically captured to record cooperation markers during measurement experiments.
		
		\subsection{Image Enhancement Under Extreme Illumination}
\begin{figure*}[t]
			\centering
			\centering
			\setlength{\abovecaptionskip}{0.cm}
			\includegraphics[scale=0.5381]{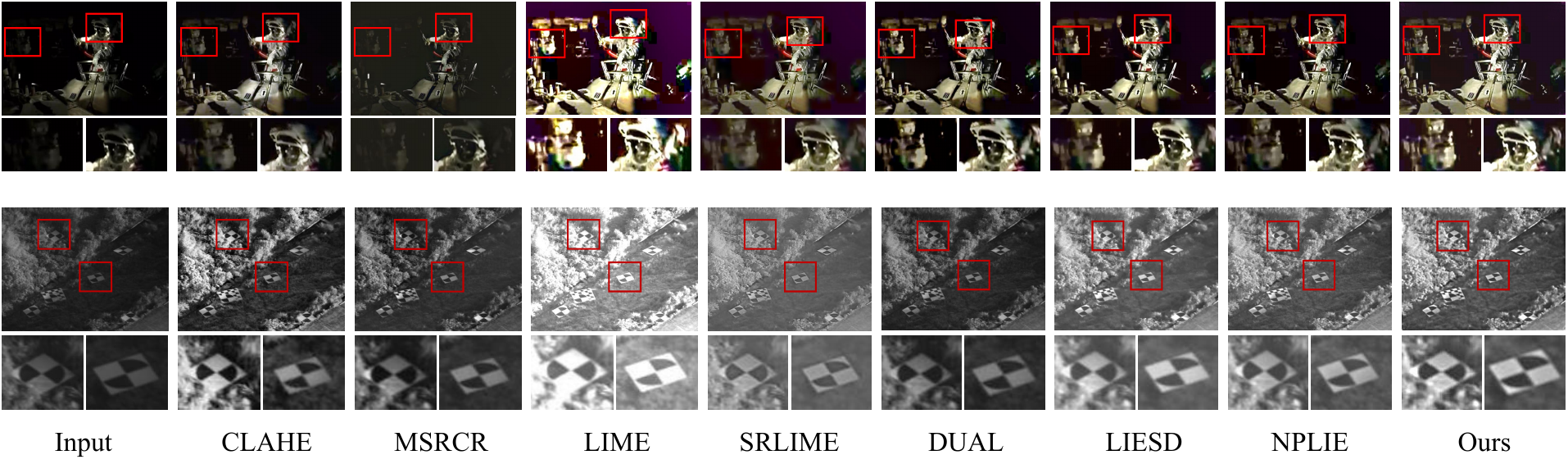}
			\caption{Comparisons in low-light condition enhancement results. The red box indicates magnified areas of detail. From left to right, each column represents input images, CLAHE \cite{1}, MSRCR \cite{2}, LIME \cite{3}, SRLIME \cite{4}, DUAL \cite{5}, LIESD \cite{6}, NPLIE \cite{7}, and our proposed method.}
			\label{fig:low-light condition}
			
			\vspace{0.5cm}
			\centering
			\centering			
			\includegraphics[scale=0.523]{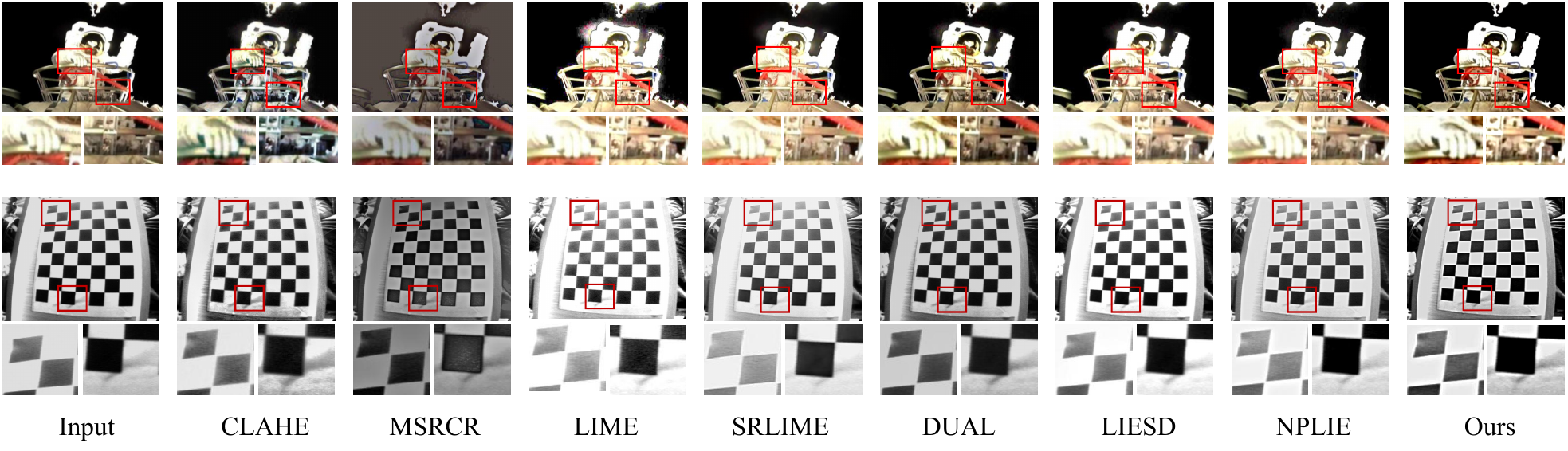}
			\caption{Comparisons in high-light condition enhancement results. The red box indicates magnified areas of detail. From left to right, each column represents input images, CLAHE \cite{1}, MSRCR \cite{2}, LIME \cite{3}, SRLIME \cite{4}, DUAL \cite{5}, LIESD \cite{6}, NPLIE \cite{7}, and our proposed method.}
			\label{fig:high-light condition}
			
			\vspace{0.5cm}
			\centering
			\centering
			\includegraphics[scale=0.5275]{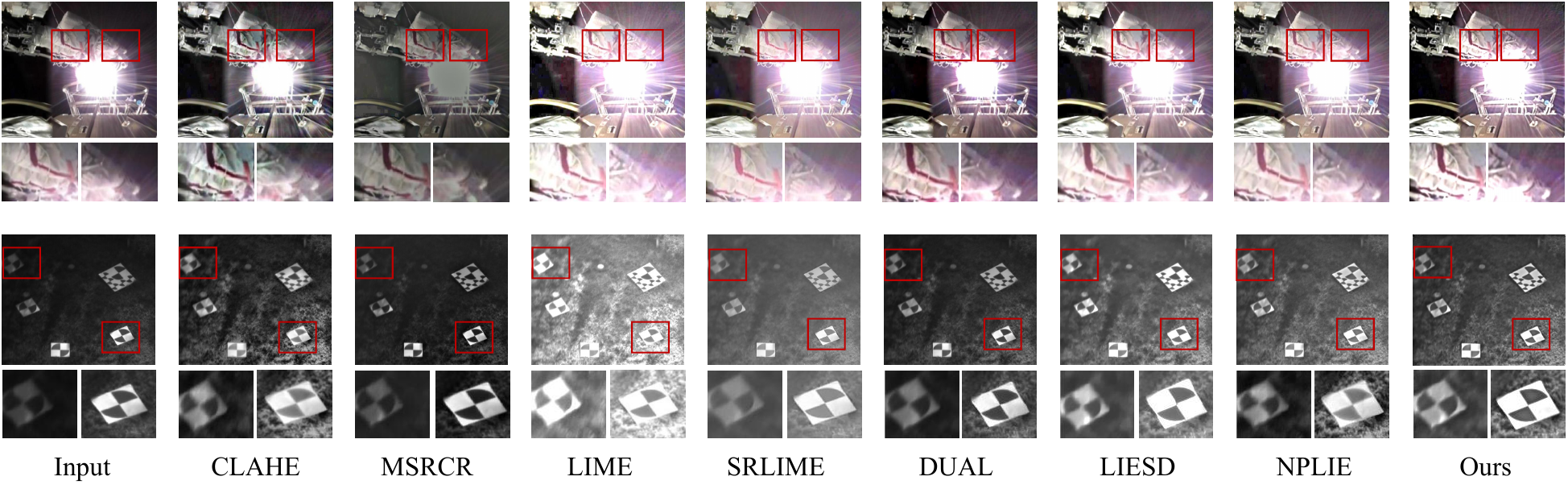}
			\caption{Comparisons in uneven-light condition enhancement results. The red box indicates magnified areas of detail. From left to right, each column represents input images, CLAHE \cite{1}, MSRCR \cite{2}, LIME \cite{3}, SRLIME \cite{4}, DUAL \cite{5}, LIESD \cite{6}, NPLIE \cite{7}, and our proposed method.}
			\label{fig:uneven-light condition}
		\end{figure*}

		\emph{1) Qualitative evaluation:}	
		To evaluate the performance of the methods under varying illumination conditions, a visual analysis is conducted in three groups: low-light conditions, high-light conditions, and uneven-light conditions.
		The corresponding results are illustrated in Fig.\ref{fig:low-light condition} to \ref{fig:uneven-light condition}.
		
		Fig.\ref{fig:low-light condition} depicts a low-light setting with numerous regions of diminished visibility. The results of the eight enhancement algorithms show that all methods effectively enhance image luminance and visibility. Nonetheless, the performances of MSRCR, LIME, and SRLIME are disappointing, characterized by excessive or insufficient corrections. Notably, CLAHE tends to excessively preserve contrast, leading to distortion. In contrast, DUAL, LIESD, and NPLIE effectively increase brightness in dark regions while maintaining image contrast. However, our method excels in detail processing, as evidenced by the enhanced clarity of the astronaut helmet and robotic arm details, alongside the natural and pronounced contrast of diagonal markers.
		
		Fig.\ref{fig:high-light condition} illustrates a scene with adequate illumination. Given that most of the existing algorithms are designed for low-light conditions, certain algorithms may yield distorted results when applied to well-lit scenes. For instance, MSRCR, LIME, and SRLIME may exhibit varying degrees of distortion and artifacts, particularly noticeable in the astronauts' head positions. LIESD and NPLIE demonstrate unsatisfactory performance in preserving the contrast of bright areas, as evidenced by the loss of details in the astronaut's finger seam. Although CLAHE exhibits superior contrast retention, it suffers from color distortion post-correction. In comparison, while DUAL exhibits favorable processing effects, our algorithm slightly surpasses it in achieving a balance between brightness and detail preservation.
		
		Fig.\ref{fig:uneven-light condition} presents a comparison under uneven-light conditions, which commonly occurs in practical scenarios. The images of diagonal markers reveal that CLAHE, LIME, SRLIME, and NPLIE exhibit subpar performance in contrast retention. Although MSRCR maintains commendable contrast, it still falls short of achieving optimal brightness balance. On the other hand, DUAL, LIESD, and our proposed algorithm demonstrate satisfactory performance in both contrast and brightness balance. However, ours outperforms the others when evaluating the texture details on the spacesuit. The outstanding performance of our algorithm under uneven illumination conditions sets the stage for subsequent enhancements in measurement accuracy and stability.		
		
		\begin{table*}[htbp]
			\setlength{\belowcaptionskip}{-0.2cm}
			\begin{center}
				\caption{Comparison of average \textbf{NIQE} \cite{8} on two datasets}
				\LARGE 
				\renewcommand\arraystretch{1.25} 
				\centering   
				\label{tab:NIQE}    
				\resizebox{0.86\textwidth}{!}{
					\begin{tabular}{c|c c c c c c c c c}
						\toprule[1.8pt]
						Method  & Input & CLAHE \cite{1} & MSRCR \cite{2}& LIME \cite{3} & SRLIME \cite{4} & DUAL \cite{5} & LIESD\cite{6} & NPLIE \cite{7} & Ours \bigstrut\\
						\hline
						\textsl{Data1}   & 3.98  & 3.4   & 3.43  & 3.62 & \textbf{3.27}  & 3.41 & 3.44  & 3.56  & \underline{3.35} \bigstrut\\
						\textsl{Data2}   & 4.99  & 4.58  & \underline{4.05}  & 4.89 & 4.44   & 4.43 & 4.48  & 4.29  & \textbf{4.11} \\
						\rowcolor{gray!15} Average & 4.49  & 3.99  & \underline{3.74}  & 4.26 & 3.86   & 3.92 & 3.96  & 3.93  & \textbf{3.73} \bigstrut\\
						\bottomrule[1.8pt]
						\multicolumn{4}{l}{\Large The best results are bold, and the second best results are underlined.}\\
				\end{tabular} }
			\end{center}
			
			\vspace{0.15em}
			\begin{center}
				\caption{Comparison of average \textbf{ARISM} (luminance only) \cite{9} on two datasets}
				\LARGE 
				\renewcommand\arraystretch{1.25} 
				\centering     
				\label{tab:ARISM-1}   
				\resizebox{0.86\textwidth}{!}{
					\begin{tabular}{c|c c c c c c c c c}
						\toprule[1.8pt]
						Method  & Input & CLAHE \cite{1} & MSRCR \cite{2}& LIME \cite{3} & SRLIME \cite{4} & DUAL \cite{5} & LIESD\cite{6} & NPLIE \cite{7} & Ours  \bigstrut\\
						\hline
						\textsl{Data1}   & 1.46  & 1.23  & \textbf{1.18} & 1.24 & 1.24   & 1.24 &1.22  & 1.23  & \underline{1.21} \bigstrut\\
						\textsl{Data2}   & 1.61  & 1.20 & 1.23 & 1.22 & 1.23  & \textbf{1.15} & 1.21  & 1.24  & \underline{1.17} \\
						\rowcolor{gray!15} Average & 1.54  & 1.22  & 1.21  & 1.23 & 1.24   &\underline{1.20} & 1.22 & 1.24  & \textbf{1.19} \bigstrut\\
						\toprule[1.8pt]
						\multicolumn{4}{l}{\Large The best results are bold, and the second best results are underlined.}\\
				\end{tabular} } 
			\end{center}
			
			\vspace{0.15em}
			\begin{center}
				\caption{Comparison of average \textbf{ARISM} (luminance and chromatic) \cite{9} on two datasets}
				\LARGE 
				\renewcommand\arraystretch{1.25} 
				\centering     
				\label{tab:ARISM-2}    
				\resizebox{0.86\textwidth}{!}{ 
					\begin{tabular}{c|c c c c c c c c c}
						\toprule[1.8pt]
						Method  & Input & CLAHE \cite{1} & MSRCR \cite{2}& LIME \cite{3} & SRLIME \cite{4} & DUAL \cite{5} & LIESD\cite{6} & NPLIE \cite{7} & Ours \bigstrut\\
						\hline
						\textsl{Data1}   & 1.68  & 1.14  & \textbf{1.10} & 1.18 & 1.18 & 1.15 &1.14  & 1.15  & \underline{1.11} \bigstrut\\
						\textsl{Data2}   & 1.84  & 1.11 & \textbf{1.05} & 1.14 & 1.15  & \underline{1.07} & 1.13  & 1.16 & 1.08 \\
						\rowcolor{gray!15} Average & 1.76  & 1.13  & \textbf{1.08}  & 1.16 & 1.17   &1.11 & 1.14 & 1.16  & \underline{1.10} \bigstrut\\
						\toprule[1.8pt]
						\multicolumn{4}{l}{\Large The best results are bold, and the second best results are underlined.}\\
				\end{tabular} } 
			\end{center}
			
			\vspace{0.15em}
			\begin{center}
				\caption{Comparison of average \textbf{NIQMC} \cite{10} on two datasets}
				\LARGE   
				\renewcommand\arraystretch{1.25} 
				\centering     
				\label{tab:NIQMC}   
				\resizebox{0.86\textwidth}{!}{
					\begin{tabular}{c|c c c c c c c c c}
						\toprule[1.8pt]
						Method  & Input & CLAHE \cite{1} & MSRCR \cite{2}& LIME \cite{3} & SRLIME \cite{4} & DUAL \cite{5} & LIESD\cite{6} & NPLIE \cite{7} & Ours  \bigstrut\\
						\hline
						\textsl{Data1}   & 4.09  & 4.44  & 4.39 & 4.99 & 5.21  &\underline{5.27}  & 5.17  & 5.22  & \textbf{5.28} \bigstrut\\
						\textsl{Data2}   & 4.14  & 4.14 & 4.27 & 4.56 & 4.31  & 4.48 & 4.68  &\underline{4.85} & \textbf{4.95} \\
						\rowcolor{gray!15} Average & 4.12  & 4.29  & 4.33  & 4.78 & 4.76   &4.88 & 4.93 & \underline{5.04}  & \textbf{5.12}\bigstrut\\
						\toprule[1.8pt]
						\multicolumn{4}{l}{\Large The best results are bold, and the second best results are underlined.}\\
				\end{tabular} } 
			\end{center}
		\end{table*}
		
		\begin{figure*}[t]
			\centering
			\centering
			\setlength{\abovecaptionskip}{0.cm}
			\includegraphics[scale=0.519]{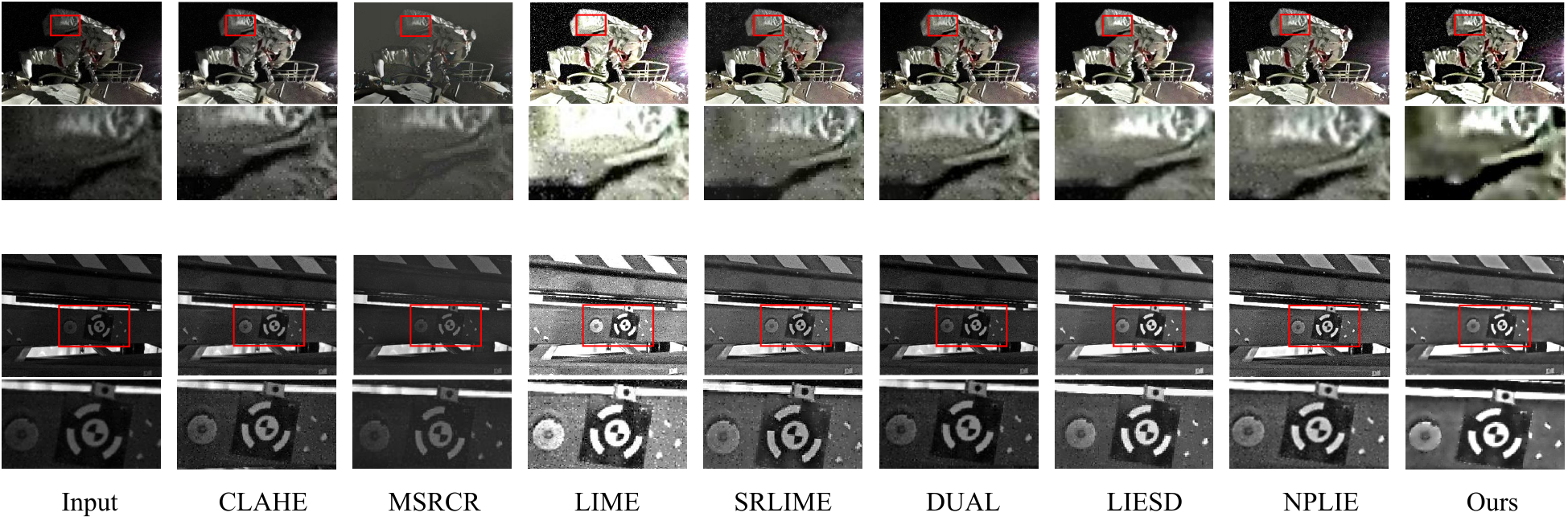}
			\caption{Comparison of different methods of denoising. The red box indicates magnified areas of detail. From left to right, each column represents input images, CLAHE \cite{1}, MSRCR \cite{2}, LIME \cite{3}, SRLIME \cite{4}, DUAL \cite{5}, LIESD \cite{6}, NPLIE \cite{7}, and our proposed method.}
			\label{fig:noise_img}
		\end{figure*}
		
		\emph{2) Quantitative evaluation:}
		In addition to the qualitative comparisons discussed earlier, quantitative metrics are employed to assess and compare the proposed method against others. Since ground-truth reference images are not available for the test images shown in Fig.\ref{fig:low-light condition} to \ref{fig:uneven-light condition}, the natural image quality evaluator (NIQE) \cite{8}, autoregressive-based image sharpness metric (ARISM) \cite{9}, and no-reference image quality metric for contrast distortion (NIQMC) \cite{10} are employed for assessment. The quantitative scores obtained are summarized in TABLE \ref{tab:NIQE} to \ref{tab:NIQMC}. In these tables, the best results are bold, and the second best results are underlined.
		
		The NIQE index is used to evaluate image quality, where lower scores indicate better-perceived image quality. As shown in TABLE \ref{tab:NIQE}, our proposed method achieves the overall lowest score, indicating clearer and less distorted results.
		The ARISM indicator has two variants: one that evaluates only luminance and another that evaluates both luminance and chromatic components. Smaller values of ARISM denote better performance. In TABLE \ref{tab:ARISM-1} and TABLE \ref{tab:ARISM-2}, the scores of MSRCR and DUAL are competitive with those of the proposed method. However, based on the aforementioned visual evaluation, our proposed method exhibits superior visual effects, yielding higher comprehensive scores.
		For NIQMC, larger values indicate better image contrast quality. As observed in TABLE \ref{tab:NIQMC}, our proposed method achieves the highest score, signifying successful enhancement of overall image quality without introducing excessive artifacts.
		Overall, our proposed method demonstrates favorable results in terms of quantitative evaluation indicators.
		
		\begin{figure}[t]
			\centering
			\includegraphics[scale=0.45]{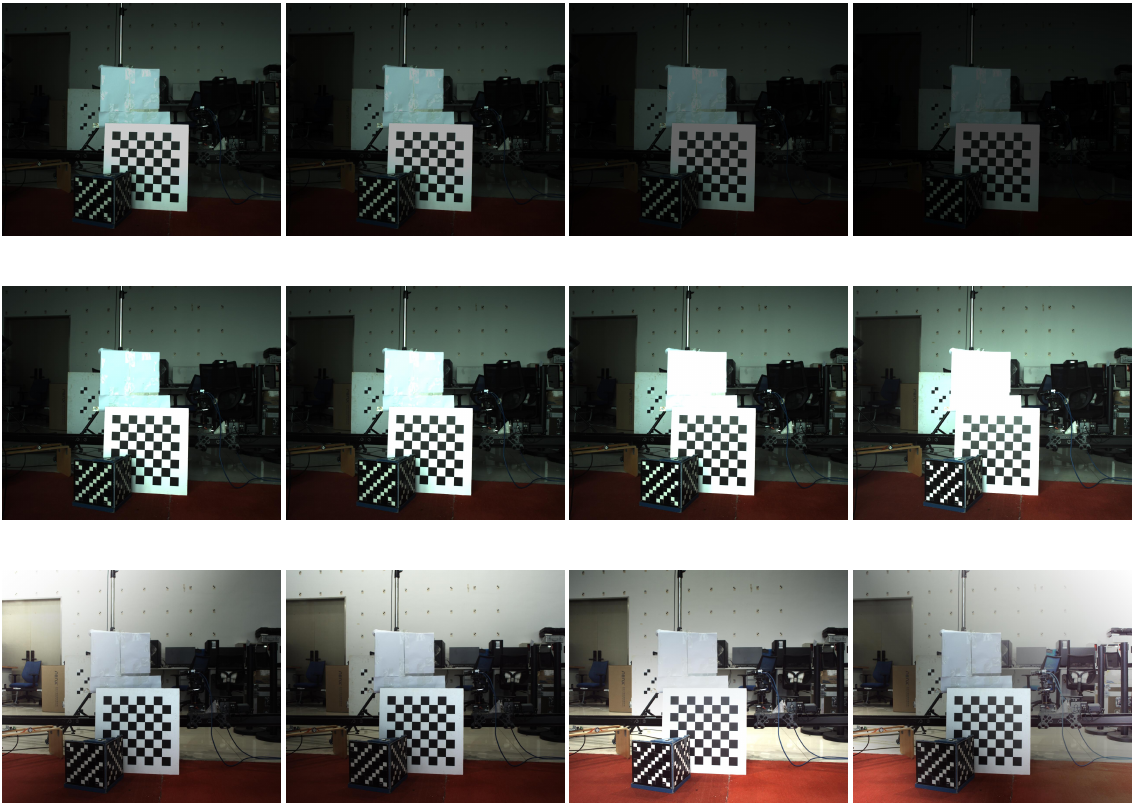}
			\centering
			\caption{Positioning accuracy experiments under different illumination conditions: the first column represents the low-light condition with decreasing illumination; the second column represents the high-light condition with increasing illumination; the third column represents various cases of uneven light.}
			\label{fig:Positioning accuracy}
		\end{figure}
		\begin{table*}[t]
			\begin{center} 
				\caption{Average \textbf{PSNR} and \textbf{SSIM} result of different enhancement methods}
				\normalsize     
				\renewcommand\arraystretch{1.25} 
				\centering     
				\label{tab:noise}   
				\scalebox{0.82}
				{
					\begin{tabular}{c c c c c c c c c}
						\toprule[1.5pt]
						\quad  & CLAHE \cite{1} & MSRC \cite{2} & LIME \cite{3} & SRLIME \cite{4} & DUAL \cite{5}  & LIESD \cite{6} & NPLIE \cite{7}  & Ours \bigstrut\\                   
						\hline
						SSIM   & 0.62  & 0.43 & 0.56 & 0.75 & 0.74  &0.73  & \underline{0.77} & \textbf{0.81} \bigstrut\\
						PNSR   & 30.87  & 30.59 & 20.97& 25.20 & \underline{29.86}  & 25.20 & 29.75  & \textbf{32.15} \bigstrut\\
						\toprule[1.5pt]
						\multicolumn{4}{l}{\small The best results are bold, and the second best results are underlined.}\\
				\end{tabular} }
			\end{center}
			\vspace{-0.25cm}
		\end{table*}
		
		\subsection{Noise Suppression}
		This section validates the noise suppression capability of our proposed algorithm,  demonstrating its effectiveness in enhancing complex low-light images with substantial noise, as shown in Fig.\ref{fig:noise_img}. The figure illustrates that existing methods tend to amplify the considerably strong noise in regions with very low light intensity. While MSRCR, LIME, DUAL, and NPLIE enhance the visibility of complex low-light images to some extent, their results still display noticeable noise artifacts. CLAHE and SRLIMR, despite considering noise reduction, struggle to effectively handle noise near the edges.
		Comparatively, LIESD produces a slightly better denoising effect but sacrifices image details and leaves residual noise at the edges. In contrast, our proposed method achieves superior edge preservation and global contrast enhancement, resulting in sharper and visually appealing outcomes.
		
		To quantitatively evaluate the effectiveness of our proposed method, a comparison is conducted with a competing method (followed by post-processing using the denoising method BM3D \cite{BM3D}) on a dataset of 100 images mentioned above. For the synthesis of low-light images, gamma correction (with $\gamma $ = 2.5) is first applied to the acquired images. Subsequently, Poisson noise and White Gaussian noise are added to simulate typical noise observed in natural images \cite{4}.
		The peak signal-to-noise ratio (PSNR) and structural similarity index measure (SSIM) are used to analyze the results of noise suppression. PSNR is the most widely used objective image evaluation metric based on the error between corresponding pixels. SSIM assesses image similarity from three aspects: luminance, contrast, and structure, outputting values within the range of 0 to 1, with higher scores indicating less image distortion.
		The average PSNR and SSIM values of the images processed by the compared methods are presented in TABLE \ref{tab:noise}. It is observed that the proposed method achieves the highest PSNR and SSIM values among the compared methods, even surpassing those obtained by methods followed by post-processing.
		
		\begin{figure}[t]
			\centering
			\includegraphics[scale=0.41]{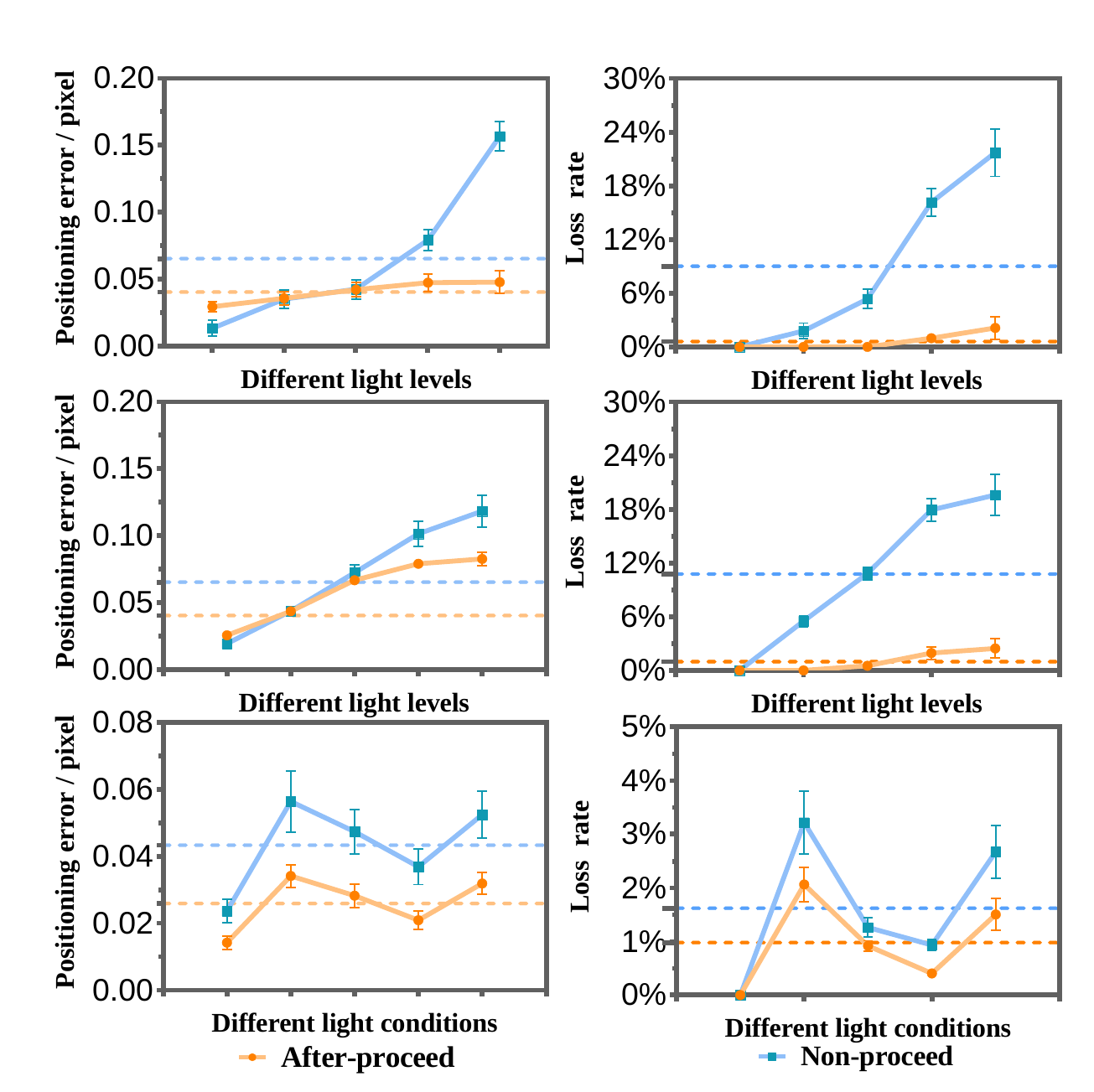}
			\centering
			\caption{Location results under different illumination conditions. From top to bottom respectively: low-light condition, high-light condition, and uneven-light condition.}
			\label{fig:Location results}
			\vspace{-0.15cm}
		\end{figure}
		
		\subsection{Feature Extraction Application}
		The precision of measurements in practical photogrammetric applications often depends on the quality of the image. In this section, a series of experiments are conducted to evaluate the influence of our proposed method on feature extraction accuracy. Ground truth values are established using the positions of diagonal markers under optimal illumination conditions. Subsequently, various extreme scenarios are simulated by manipulating the illumination conditions. Marker extraction is performed using the template matching method \cite{Automatic,Matching}, and the experiments are divided into three groups based on different illumination conditions: low-light, high-light, and uneven-light, as shown in Fig.\ref{fig:Positioning accuracy}.
		
		The influence of various illumination conditions on marker positioning accuracy is depicted in Fig.\ref{fig:Location results}, corresponding to Fig.\ref{fig:Positioning accuracy}. As the degree of illumination variation increases, there is a corresponding rise in localization error, potentially leading to a complete loss of localization under extreme conditions. Nevertheless, it is worth noting that regardless of illumination conditions, the processed images consistently exhibit improved positioning accuracy, resulting in a significant reduction in feature loss rates. 
		Interestingly, optimal lighting conditions reveal heightened positioning errors in the processed images, primarily attributed to the template matching approach employed in the positioning methodology. Template matching relies on template-pixel correlation to determine the optimal match position. However, integration of enhancement algorithms introduces nonlinear transformations to the grayscale information of the target, influencing match correlation adversely \cite{Automatic,Matching,Guan_TCYB2021}. Nevertheless, as lighting conditions become more extreme, this impact diminishes relative to environmental interferences, resulting in notable improvements in positioning accuracy post-processing. Moreover, this challenge could be addressed through the integration of prediction mechanisms or improvements in the matching method. 
		
		\begin{figure}[t]
			\centering		
			\subfloat[]
			{\includegraphics[scale=0.293]{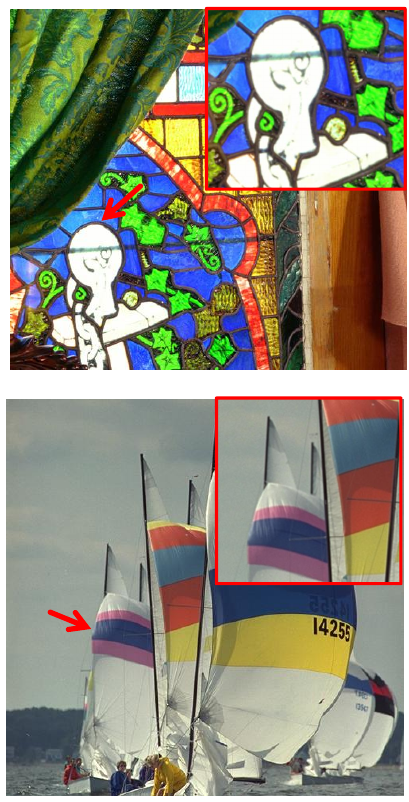}}
			\hspace{0.005cm}
			\subfloat[]	
			{\includegraphics[scale=0.293]{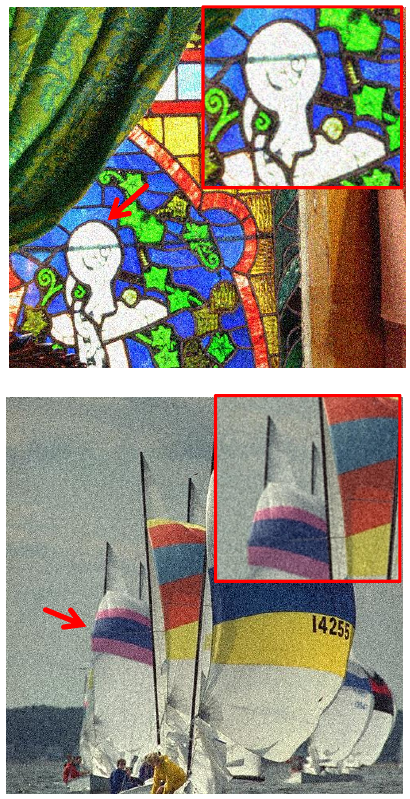}}
			\hspace{0.005cm}	
			\subfloat[]	
			{\includegraphics[scale=0.293]{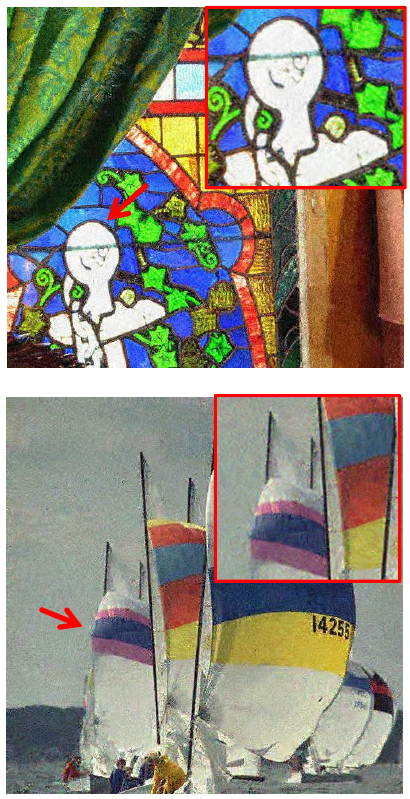}}
			\hspace{0.005cm}	
			\subfloat[]	
			{\includegraphics[scale=0.293]{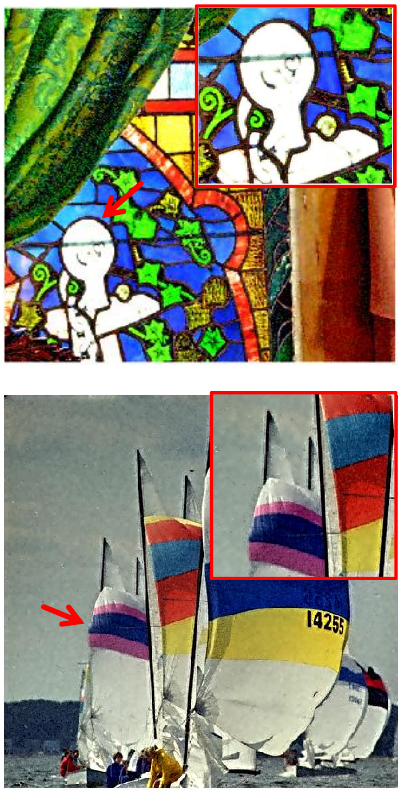}}
			\caption{Comparison of noise suppression results. (a) Source images. (b) Noise images (Gaussian noise: $\mu $= 0, $\sigma $= 0.02). (c) Results by \cite{BM3D}. (d) Results by our framework. }
			\label{fig:others-1}
			
			\vspace{0.2cm}
			
			\centering		
			\subfloat[]
			{\includegraphics[scale=0.3]{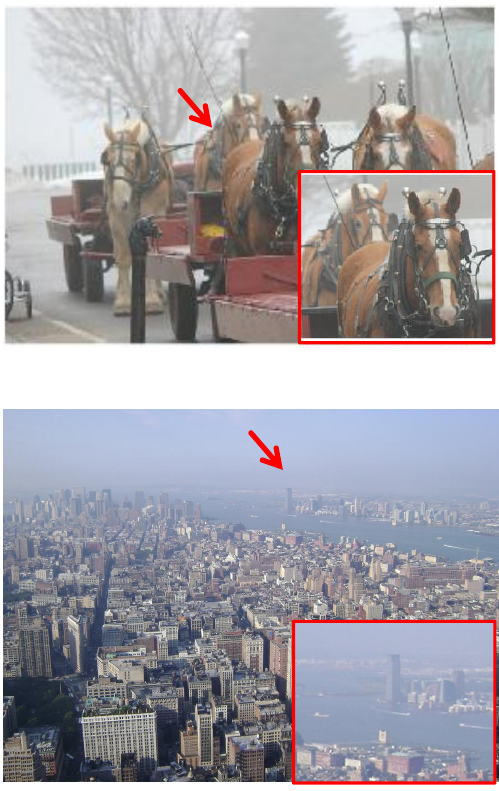}}
			\hspace{0.08cm}
			\subfloat[]	
			{\includegraphics[scale=0.3]{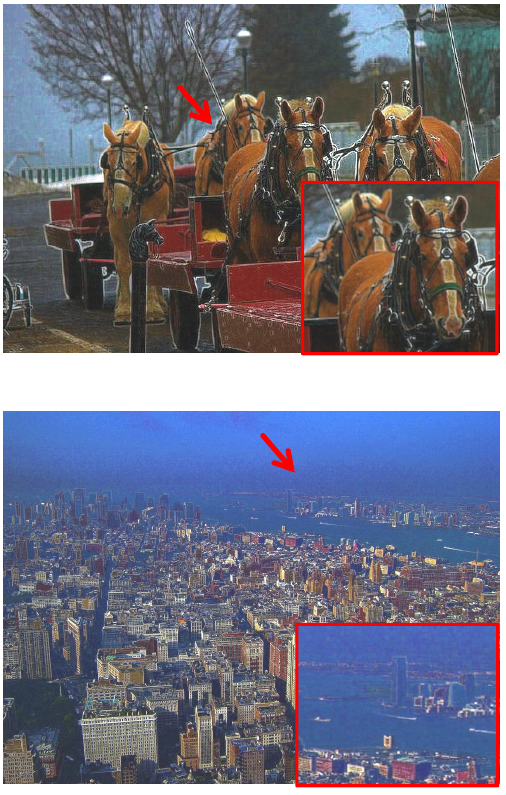}}
			\hspace{0.08cm}	
			\subfloat[]	
			{\includegraphics[scale=0.3]{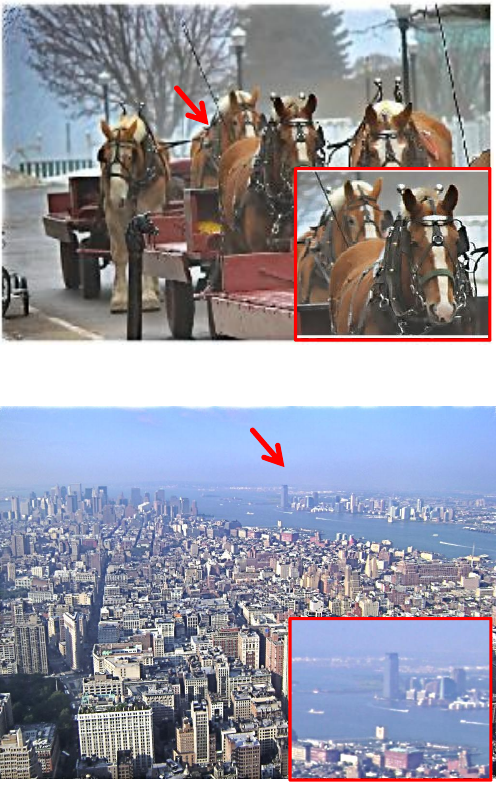}}
			\caption{Comparison of haze removal results. (a) Haze images. (b) Results by \cite{dark}. (c) Results by our framework. }
			\label{fig:others-2}
		\end{figure}	
		
		\subsection{Other Application}
		Furthermore, our proposed method provides versatile applications like noise suppression and image haze removal. In this section, we use publicly available common datasets for processing.
		
		\vspace{0.1cm}
		\emph{1) Noise suppression:} The improved guided filter used in the framework adeptly suppresses noise. Fig.\ref{fig:others-1} presents comparisons of noise suppression results based on our framework and the denoising method based on block-wise estimate \cite{BM3D}. It can be seen that our framework effectively reduces noise while preserving details in daily scenes. Furthermore, our method effectively avoids excessive brightness correction in scenes with balanced illumination.
		
		\vspace{0.1cm}
		\emph{2) Haze removal:} Inspired by \cite{3} and \cite{6}, we regard the inverted haze image as a low-light image. Thus, the dehazed image can be obtained by inputting the inverted haze image into our framework and subsequently inverting the output image again. Fig.\ref{fig:others-2} presents comparisons of haze removal results based on our framework and the dehazing method based on dark channel prior \cite{dark}. Our method effectively mitigates the majority of the haze while preserving the natural visual appearance.
		
		\section{Conclusion}
		This paper introduces a novel framework aimed at addressing the challenges of enhancing image contrast and suppressing noise amplification under extreme conditions. The proposed framework utilizes the gradient-domain weighted guided filter (GDWGIF) to effectively estimate illumination and enhance image details, resulting in superior results. By concurrently processing the illumination and reflection layers, our framework achieves improved performance and visually pleasing results. Extensive experimentation validates significant enhancements in contrast, brightness, and preservation of image details. Quantitative evaluation metrics further support the superiority of our proposed framework compared to alternative approaches, thereby benefiting subsequent image measurement tasks.
		
		\bibliographystyle{IEEEtran}
		\bibliography{ref}
		\newpage
		
	\end{CJK}	
\end{document}